%% file: main.tex
\documentclass[10pt,twocolumn,letterpaper]{article}

\usepackage{cvpr}              

\usepackage{url}
\usepackage[utf8]{inputenc} 
\usepackage[T1]{fontenc}    
\usepackage{booktabs}       
\usepackage{amsfonts}       
\usepackage{nicefrac}       
\usepackage{microtype}      
\usepackage{xcolor}         
\usepackage{multirow}
\usepackage{multicol}
\usepackage{wrapfig}
\usepackage{mathtools}
\usepackage{enumitem}
\usepackage{arydshln}
\usepackage{bm}
\usepackage{booktabs}
\usepackage{subcaption}
\usepackage{balance}
\usepackage{appendix}

\input{preamble}
\definecolor{cvprblue}{rgb}{0.21,0.49,0.74}
\usepackage[pagebackref,breaklinks,colorlinks,allcolors=cvprblue]{hyperref}

\def\paperID{7594} 
\def\confName{CVPR}
\def\confYear{2026}

\newcommand{\NAME}{HumanNOVA}

\title{\NAME: Photorealistic, Universal and Rapid  \\ 3D Human Avatar Modeling from a Single Image}

\author{
  {Hezhen Hu}$^{1}$ \quad
  {Wangbo Zhao}$^{2}$ \quad
  {Lanqing Guo}$^{1}$ \quad
  {Hanwen Jiang}$^{1}$ \quad
  {Jonathan C. Liu}$^{1}$ \quad \\
  {Zhiwen Fan}$^{3}$ \quad
  {Kai Wang}$^{2}$ \quad
  {Zhangyang Wang}$^{1}$ \quad
  {Georgios Pavlakos}$^{1}$ \\
  $^{1}$ University of Texas at Austin \quad
  $^{2}$ National University of Singapore \quad
  $^{3}$ Texas A\&M University \\
}

\begin{document}

\twocolumn[{%
\renewcommand\twocolumn[1][]{#1}%
\maketitle
\begin{center}
    \newcommand{\teaserwidth}{\textwidth}
    \vspace{-2.0em}
    \centerline{
    \includegraphics[width=.99\linewidth]{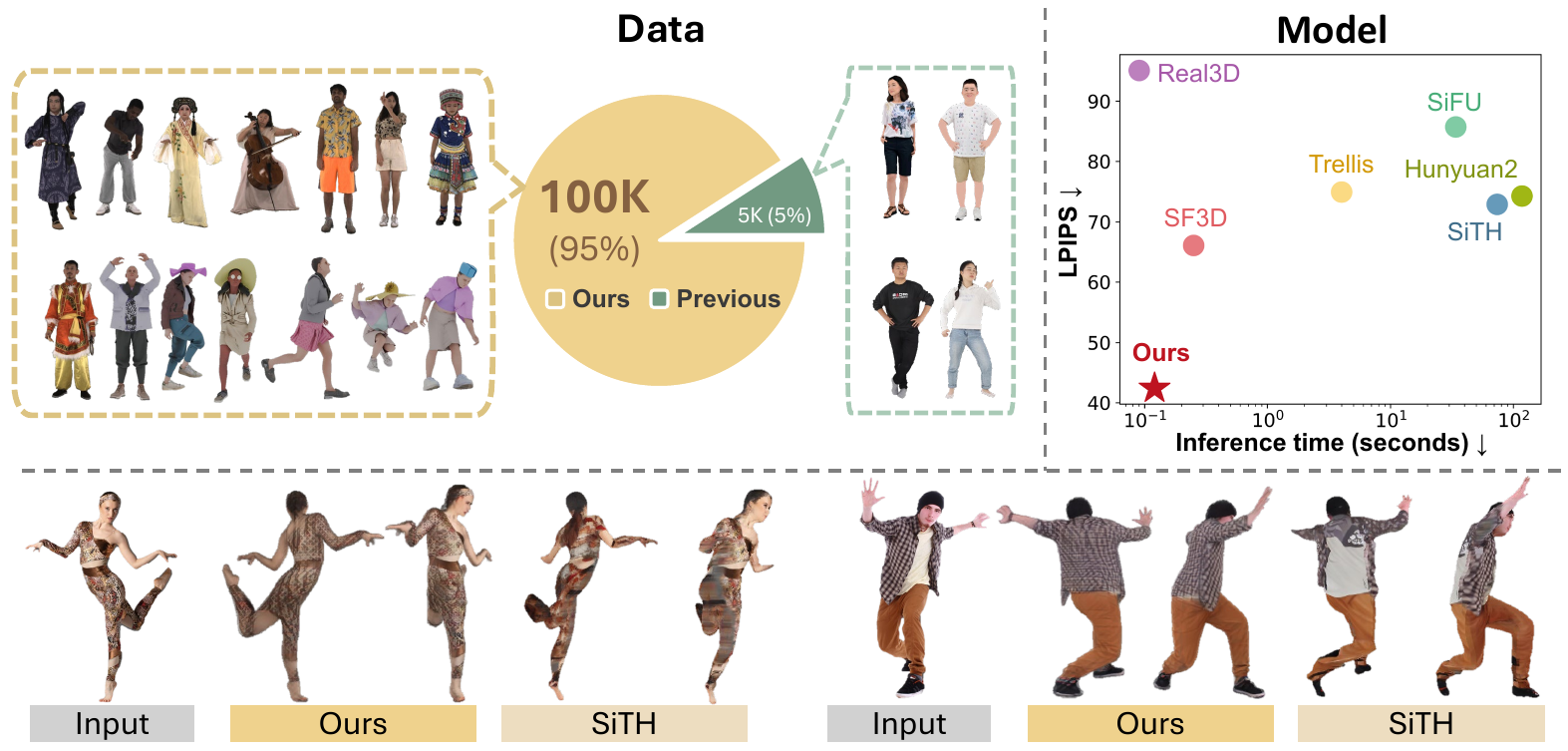}
     }
\vspace{-0.8em}
   \captionof{figure}{\textbf{Photorealistic, universal and rapid 3D human avatar modeling from a single image by the proposed approach, \NAME.}    
    It benefits from both our generated large-scale data and feed-forward model design. Our data generation pipeline expands training data by 20 times (top-left for visualization). With this data, \NAME~achieves superior performance while maintaining \emph{rapid} inference among existing methods (top-right). Once trained, it is \emph{universal} without the need for test-time fine-tuning or adaptation. Qualitative results show that \NAME~produces more precise \emph{photorealistic} reconstructions compared to the state-of-the-art SiTH method~\cite{ho2024sith} (bottom).}
   \vspace{-1.3em}
\label{fig:teaser}
\end{center}%
}]

\input{sec/0_abstract}    
\input{sec/1_intro}
\input{sec/2_related_work}
\input{sec/3_HumanNOVA}

\input{sec/4_experiments}

\input{sec/5_conclusion}

\balance
{
    \small
    \bibliographystyle{ieeenat_fullname}
    \bibliography{main}
}

\clearpage

{\noindent\large\bfseries Supplementary Material\par}
\vspace{0.8em}

\nobalance
\input{sec/6_supp}



\end{document}

%% file: sec/0_abstract.tex
\begin{abstract}
In this paper, we present \NAME, a photorealistic, universal, and rapid model for generating 3D human avatars from a single RGB image.
Achieving both photorealism and generalization is challenging due to the scarcity of diverse, high-quality 3D human data.
To address this, we build a scalable data generation pipeline that follows two strategies.
The first one is to leverage existing rigged assets and animate them with extensive poses from daily life.
The second strategy is to utilize existing multi-camera captures of humans and employ fitting to generate more diverse views for training.
These two strategies enable us to scale up to 100k assets, significantly enhancing both the quantity and the diversity of data for robust model training.
In terms of the architecture,  \NAME{} adopts a feed-forward, token-conditioned avatar modeling framework that allows fast inference in less than one second and requires no test-time optimization.
Given an input image and an estimated simplified human mesh (SMPL) without detailed geometry or appearance, the model first encodes both inputs into compact token representations.
These tokens then act as conditioning signals and are fused through cross-attention to construct a triplane-based 3D avatar representation.
Extensive experiments on multiple benchmarks demonstrate the superiority of our approach, both quantitatively and qualitatively, as well as its robustness under diverse input image conditions.
Project page at \url{https://HumanNOVA.github.io}.
\end{abstract}

%% file: sec/1_intro.tex
\section{Introduction}
\label{sec:intro}
Photorealistic 3D human avatar modeling has gained significant attention due to its wide applications, including virtual reality~\cite{mikhailova2024age,xue2025infinihuman}, telepresence~\cite{tu2024tele}, and human-computer interaction~\cite{cork2024just}. 
All these applications require \textit{fast} and \textit{photorealistic} 3D human modeling, ideally from \textit{a single image}.
However, reconstructing a static 3D avatar from a single view is naturally an ill-posed problem~\cite{sinha1993recovering}, due to the ambiguity of the unseen viewpoints.
To deal with this problem, existing single-view photorealistic 3D human avatar modeling methods~\cite{ho2024sith,zhang2024sifu} have explored incorporating human-related inductive bias.
These methods first inject strong structural priors via parametric body models (\emph{e.g.,} SMPL~\cite{loper2023smpl} or SMPL-X~\cite{pavlakos2019expressive}), and overlay advanced 3D representations (\emph{e.g.,} NeRF~\cite{mildenhall2020nerf} or 3D Gaussian Splatting~\cite{kerbl20233d}) to achieve photorealistic rendering.
Simultaneously, diffusion-based priors are employed to hallucinate the unseen side and back views, further alleviating view ambiguity and refining the final geometry and appearance~\cite{ho2024sith,zhang2024sifu}.
Nevertheless, this class of methods requires slow optimization on a per-instance basis, limiting the scalability and broader applications.

In this paper, we propose \textbf{\NAME}, a si\textbf{N}gle-view 3D human avatar modeling model that is ph\textbf{O}torealistic, uni\textbf{V}ersal, and r\textbf{A}pid.
\NAME{} enables high-quality reconstruction in less than one second, as shown in Figure~\ref{fig:teaser}.
The key of \NAME{} is the adoption of recently developed large-scale single-view reconstruction models, \emph{i.e.,} LRMs~\cite{hong2023lrm}.
However, building an LRM for 3D human reconstruction is non-trivial for two reasons.
First, we have a limited amount of 3D human assets for training such models.
Unlike general-category datasets, \emph{e.g.,} Objaverse~\cite{deitke2023objaverse}, which contains 800K shape instances, the current human datasets~\cite{yu2021function4d,shen2023xavatar,han2023high} usually only include a few thousand human instances.
This scarcity of training data makes it very challenging to train an LRM for human reconstruction.
Second, the current LRM architectures are designed for reconstruction of general-category objects, without incorporating category-specific priors.
However, in specific settings of high importance, like human reconstruction, we anticipate that a human-related prior will be beneficial.

To train an effective large avatar reconstruction model, we first scale up the training data.
More specifically, we design a streamlined data generation pipeline, which introduces large-scale and high-quality human data.
This data generation pipeline incorporates both synthetic and real data.
The synthetic data benefits LRM training with its large scale, while the real data is more photorealistic and is well aligned with the data distribution in real applications.
We generate 100K human instances in total, which is about 20 times larger than the combination of all existing datasets.
This largely improves the reconstruction quality of humans with LRMs.

To build on this foundation, we adjust the LRM architecture for 3D human reconstruction by incorporating 
human-related priors in the model.
Concretely, \NAME{} leverages a parametric SMPL human mesh, as estimated by~\cite{goel2023humans}, which provides a robust and reasonably accurate, yet coarse geometric prior to guide 3D human reconstruction.

Overall, our contributions are summarized as follows:
\begin{itemize}[leftmargin=*, itemsep=0pt]
    \item We introduce \NAME, a universal large human avatar reconstruction model that can reconstruct a static 3D human from a single image in less than one second. 
    We demonstrate that both data and model are crucial for extending the success of LRMs from general object reconstruction to the human domain.
    \item We design an effective data generation pipeline, which scales to 100k assets for our training dataset.
    Our pipeline considers both synthetic and real-world data, which significantly enhances both data quantity and diversity.
    \item 
    Through extensive experiments on multiple benchmarks, we demonstrate the 
    quantitative and qualitative
    superiority of \NAME{},
    achieving over 40\% relative LPIPS 
    improvement
    compared with state-of-the-art methods.
\end{itemize}

%% file: sec/2_related_work.tex
\section{Related Work}

\subsection{Single-view 3D Human Avatar Modeling}
Single-view 3D human avatar modeling is a challenging task due to its inherent ill-posed nature.
Early methods usually leveraged the mesh produced by parametric body models as the base body shape and depicted detailed clothing via mesh offsets~\cite{alldieck2019tex2shape,alldieck2018video,alldieck2019learning} or adjustable garment templates~\cite{bhatnagar2019multi,feng2023learning,jiang2020bcnet}.
However, the constraint of the mesh topology limits the capability of these methods to model complex clothing (\eg, dresses).
Recent methods tend to utilize neural representations, \eg, NeRF, SDF, and 3DGS, given their flexibility with topology modeling~\cite{kwon2021neural,albahar2023single,corona2023structured,xiu2022icon,zheng2021pamir,chen2024generalizable,dong2025moga,he2021arch++,shin2024canonicalfusion,liao2023high,xiu2023econ,huang2024tech,chu2024gpavatar,wang2025humandreamer,yang2025sigman,wang2024geneman,lu2025gas,zhuang2024idolinstant}.
For improved texture quality, researchers refine human meshes generated through iterative optimization as a guiding signal, while simultaneously leveraging image priors from diffusion models~\cite{zhang2024global,zhan2024semantic,pan2024humansplat,li2024pshuman,ho2024sith,zhang2024sifu,xue2024human}.
For example, GTA~\cite{zhang2024global} combines encoding with a visual transformer and a decoder using cross-attention on triplane features.
DIFu~\cite{song2023difu} introduces a depth‑guided implicit function that uses front and hallucinated back depth maps projected into a 3D volume to provide voxel-level priors for better surface details. 
\cite{pan2024humansplat} proposes a framework built on 3DGS, which integrates a multi-view diffusion model and a latent reconstruction transformer, to recreate generalizable human views.
SiTH~\cite{ho2024sith} decomposes this task into hallucination of images from unseen views with diffusion models and reconstruction by leveraging skinned body meshes.
\cite{weng2024template}
designs a multi-stage framework to leverage the prior from large reconstruction and diffusion models.
SIFU~\cite{zhang2024sifu} employs a side-view transformer coupled with the fitted SMPL-X mesh to map $2$D features to $3$D space, followed by the incorporation of priors via a diffusion model. 
The limitation of these approaches is the excessive reliance on diffusion priors.
This results in significantly increased inference time and might lead to restricted generalization capabilities when the training data is limited.
In this work, we 
adopt 
a different perspective by developing a fast inference framework while significantly scaling up our training data.

\subsection{Single-view General Object Modeling}
Early work for 3D object modeling from a single view focused on category-specific approaches~\cite{kanazawa2018learning, tatarchenko2019single, xu2019disn}.
Recently, researchers have explored building models that enable 3D reconstruction of general category objects from a single view.
One group of approaches finetunes generative models to generate novel views and reduces the problem to multi-view reconstruction.
A pioneering work in this space is Zero-1-to-3~\cite{liu2023zero}, which utilizes a viewpoint-conditioned diffusion model trained on synthetic assets.
Follow-up works focus on tackling the potential inconsistencies in the generated multi-view images~\cite{long2024wonder3d, shi2023mvdream, liu2023syncdreamer}.
Even without inconsistencies, the final 3D representation needs to be derived by optimization given the generated views, which can be computationally expensive.
Another line of work is building large-scale feed-forward models to directly learn 3D reconstruction from data~\cite{hong2023lrm, jiang2023leap, boss2024sf3d, zou2024triplane, wei2024meshlrm, szymanowicz2024splatter}.
LRM~\cite{hong2023lrm} is one of the pioneering works that uses a scalable transformer-based network and trains on large-scale 3D data to directly map the 2D image information to the 3D representation.
Follow-up works improve the reconstruction quality by hallucinating multiple novel views~\cite{wang2025crm, xu2024instantmesh, tang2024lgm, xu2024grm}, designing better architectures~\cite{boss2024sf3d} or curating data of larger scale~\cite{jiang2024real3d}.
To improve the generalization capabilities, Real3D~\cite{jiang2024real3d} designs a self-training procedure to augment the training data with large-scale single-view images.
SF3D~\cite{boss2024sf3d} further improves the network architecture, reducing memory consumption while enhancing the network's representational capacity for high-quality reconstruction.
In this work, we design our framework based on LRMs and explore strategies to achieve high-quality 
3D reconstruction specifically
for humans.

%% file: sec/3_HumanNOVA.tex
\section{\NAME{}}

\begin{figure*}[t!]
    \centering
    \includegraphics[width=\linewidth]{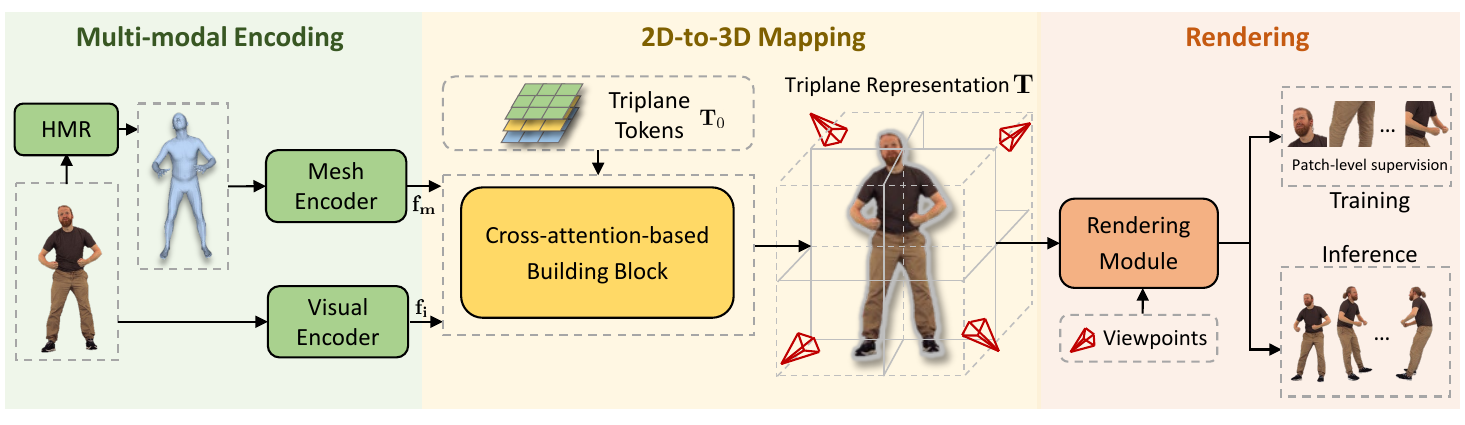}
    \vspace{-1.5em}
    \caption{\textbf{\NAME~network architecture}. Given a real-world input image, we first estimate its corresponding simplified human mesh. Image and mesh are fed into the multi-modal encoder to extract features which are utilized as the condition for the following mapping network. After that, a Transformer-based mapping network directly maps the features to the 3D triplane representation. From this triplane representation, our framework can render the 2D image given a camera viewpoint.}
    \label{fig:arch}
    \vspace{-1.em}
\end{figure*}

In this section, we first introduce the model architecture of \NAME{}, and then present our data generation method for synthesizing large-scale human training data.

\subsection{Framework}
As shown in Figure~\ref{fig:arch}, following the LRM strategy~\cite{hong2023lrm, jiang2024real3d,boss2024sf3d}, our framework achieves 3D human avatar modeling in a single feed-forward pass.
Specifically, the model consists of three main components, \ie, the multi-modal encoding, the 2D-to-3D mapping, and rendering.

Concretely, given an image $I\in \mathbb{R}^{H\times W\times 3}$, we use a DINOv2~\cite{oquab2023dinov2} encoder to tokenize it into feature tokens, denoted as $\mathbf{f_i} \in \mathbb{R}^{N_{i}\times d}$.
Here $N_i$ is the number of visual tokens, and $d$ is the latent dimension. 
Specifically, we have $N_i=HW/p^2$, where $p$ is the patch size of DINOv2.
Meanwhile, given the input image, we estimate the human surface~\cite{goel2023humans} in the form of the SMPL parametric model~\cite{loper2023smpl}, and we use PTv3~\cite{wu2024point} to tokenize the corresponding estimated human mesh.
This additional input offers an initial estimate of the body pose and body surface for the person in the image.
The output feature is denoted as $\mathbf{f_m} \in \mathbb{R}^{N_{m}\times d}$, while $N_m$ is the number of mesh tokens.

These encoded features are then fed into the mapping network to transform the image tokens $\mathbf{f_i}$ and mesh tokens $\mathbf{f_m}$ to the 3D triplane representation $\mathbf{T} \in \mathbb{R}^{3hw\times d}$, where $h$ and $w$ denote the spatial resolution of the triplane.
This triplane representation $\mathbf{T}$ is a set of learnable tokens, serving as shared initialization for all inputs.
Inspired by SF3D~\cite{boss2024sf3d}, we implement our mapping network based on PointInfinity~\cite{huang2024pointinfinity}. This mapping network consists of multiple blocks, where each block updates the triplane tokens by fusing multi-modal condition information and performing triplane refinement, which is formulated as follows:
\begin{align}
    & \mathbf{L}^l = \text{CrossAttn}(\textsc{q}=\mathbf{f_i} || \mathbf{f_m}, \textsc{kv}=\mathbf{T}^l), \\
    & \mathbf{L}^l = \text{CrossAttn}(\textsc{q}=\mathbf{L}^l, \textsc{kv}=\mathbf{f_i} || \mathbf{f_m}), \\
    &\mathbf{T}^{l+1} = \text{CrossAttn}(\textsc{q}=\mathbf{T}^l, \textsc{kv}=\mathbf{L}^l),
\end{align}
where the initial token and the final output are denoted as $\mathbf{T}_0$ and $\mathbf{T}$, respectively.
Then, we render images from the triplane following the standard ray marching method~\cite{hong2023lrm,jiang2024real3d}. We denote the process as $\hat{I}_\Phi = \pi(\mathbf{T}^*, \Phi)$, where $\pi$ denotes the rendering function, $\Phi$ is a target camera viewpoint, and $\hat{I}_\Phi$ is the image rendered under this target camera pose.

\noindent \textbf{Training objectives.}
The whole framework is jointly trained under multiple objective functions, including RGB loss $\mathcal{L}_{r}$, mask loss $\mathcal{L}_{m}$ and LPIPS loss $\mathcal{L}_{p}$ as follows:
\begin{equation}
\mathcal{L} = \frac{1}{N} \sum_{n=1}^{N} \left( \mathcal{L}_{r}^{n} + \lambda_{m} \mathcal{L}_{m}^{n} + \lambda_{p} \mathcal{L}_{p}^{n} \right),
\label{eq:loss_term}
\end{equation}
where $\lambda_{m}$ and $\lambda_{p}$ are loss weighting terms, and $N$ is the number of rendered views.
The RGB $\mathcal L_{r}$ loss and LPIPS $\mathcal L_{m}$ loss focus on the quality of the rendered image from different perspectives.
The mask loss $\mathcal L_{m}$ calculates the consistency between the accumulated density and the estimated mask.
To reduce the GPU memory consumption, all loss terms are calculated on the patch level, which is extracted by a weighted sampling strategy based on the foreground ratio of each patch to encourage the model to focus on capturing fine details and essential features.
We refer the reader to the Supplementary Material regarding the details of the patches selection implementation.

\subsection{Training Data Generation}
Next, we introduce our approach for generating large-scale synthetic data and curating high-quality real-world data.

\subsubsection{Synthetic Data Generation}
Our synthetic data generation is based on rigged human assets, where we articulate human shapes by applying real-world human poses sampled from the AMASS dataset~\cite{mahmood2019amass}. 
Here, we first introduce the preliminaries of the SMPL-X model. Then we formulate our strategy for data generation.

\noindent \textbf{Parametric body model.}
Parametric models like SMPL~\cite{loper2023smpl} and SMPL-X~\cite{pavlakos2019expressive} provide a mapping from pose and shape parameters to the mesh surface.
For example, SMPL-X
is defined as a mapping function $M(\pmb\theta, \pmb\beta, \pmb\psi): \mathbb{R}^{|\pmb\theta|} \times \mathbb{R}^{|\pmb\beta|} \times \mathbb{R}^{|\pmb\psi|} \rightarrow \mathbb{R}^{3N}$, where $\pmb\theta, \pmb\beta, \pmb\psi$ are the parameters for pose, shape, and facial expression, respectively.
The function of SMPL-X is formulated as follows:
\begin{equation}
\label{equ:mano}
  \mathbf{M}(\pmb{\beta}, \pmb{\theta}, \pmb{\psi}) = W(\mathbf{T}(\pmb{\beta}, \pmb{\theta}, \pmb{\psi}), J(\pmb{\beta}), \pmb{\theta}, \mathbf{W}),
\end{equation}
\begin{equation}
\label{equ:mano2}
  \mathbf{T}(\pmb{\beta}, \pmb{\theta}, \pmb{\psi}) = \bar{\mathbf{T}}_m + B_S(\pmb{\beta}) + B_E(\pmb{\psi}) + B_P(\pmb{\theta}),
\end{equation}
where $B_P(\cdot)$, $B_S(\cdot)$ and $B_E(\cdot)$ denote pose, shape, and expression blend functions, respectively, while $\mathbf{W}$ is the associated set of blend weights. 
The corrective blend shapes, \ie $B_P(\pmb{\theta})$,  $B_E(\pmb{\psi})$ and $B_S(\pmb{\beta})$, supply per-vertex offsets that deform the template human mesh $\bar{\mathbf{T}}_m$.
Subsequently, linear blend skinning $W(\cdot)$ rotates these deformed vertices around the joints $J(\pmb{\beta})$ and smooths the transformations via blend weights $\mathbf{W}$, returning the final human mesh.

\noindent \textbf{Synthetic data generation procedure.} During synthetic data generation, we first randomly sample a set of SMPL-X parameters from the AMASS~\cite{mahmood2019amass} dataset $\{\mathbf{R}^{\text{src}}, \mathbf{T}_{\text{src}}\} \sim \mathcal{D}_{\text{AMASS}}$.
After that, we utilize the above parameters to animate ($A(\cdot)$) the rigged human asset $\mathbf{M}$, re-center $RC(\cdot)$ the animation results, and assign a set of camera viewpoints $\mathbf{C}$ for rendering multiple views to serve as training data:
\begin{align}
\{\mathbf{I}_i\}_{i=1}^{N} &= \text{Render}(\text{RC}(A(\mathbf{M}, \{\mathbf{R}^{\text{src}}, \mathbf{T}_{\text{src}}\})), \{\mathbf{C}_i\}_{i=1}^{N}).
\end{align}

\begin{figure*}[t!]
\centering
\begin{minipage}[t]{0.30\linewidth}
    \small
    \centering
    \setlength{\tabcolsep}{2pt}
    \captionof{table}{\label{tab:ablation1} {Ablation on the generated data type on the CustomHuman dataset.}}
    \resizebox{1.0\linewidth}{!}{
    \begin{tabular}{l ccc}
    \toprule
    \textbf{Method} & {\bf PSNR}$\uparrow$ & {\bf SSIM}$\uparrow$ & {\bf LPIPS}$\downarrow$ \\ \midrule
    {w/o gen-data (assets)} & 21.84 & 0.9333 & 46.51 \\
    {w/o gen-data (multi-cam)} & 21.76 & 0.9326 & 47.83 \\
    \midrule
    {\NAME} & \textbf{22.07} & \textbf{0.9344} & \textbf{45.18}  \\
    \bottomrule
    \end{tabular}
}
\end{minipage}\hfill
\begin{minipage}[t]{0.32\linewidth}
    \small
    \centering
    \captionof{table}{\label{tab:ablation2} {Ablation on the generated data scale on the CustomHuman dataset.} }
    \resizebox{0.9\linewidth}{!}{
    \begin{tabular}{l ccc}
    \toprule
    \textbf{Ratio} & {\bf PSNR}$\uparrow$ & {\bf SSIM}$\uparrow$ & {\bf LPIPS}$\downarrow$ \\ \midrule
    {25\%} & 21.98 & 0.9313 & 50.14 \\
    {50\%} & 22.02 & 0.9338 & 47.03  \\
    \midrule
    {\NAME} & \textbf{22.07} & \textbf{0.9344} & \textbf{45.18}  \\
    \bottomrule
    \end{tabular}
}
\end{minipage}\hfill
\begin{minipage}[t]{0.32\linewidth}
    \small
    \centering
    \setlength{\tabcolsep}{4pt}
    \captionof{table}{\label{tab:ablation3} {Ablation on the impact of model settings on the CustomHuman dataset.} }
    \resizebox{1.0\linewidth}{!}{
    \begin{tabular}{l ccc}
    \toprule
    \textbf{Method} & {\bf PSNR}$\uparrow$ & {\bf SSIM}$\uparrow$ & {\bf LPIPS}$\downarrow$ \\ \midrule
    {w/o mesh prior} & 21.89 & 0.9334 & 46.26 \\
    {small triplane size (32)} & 21.78 & 0.9323 & 48.33  \\
    \midrule
    {\NAME} & \textbf{22.07} & \textbf{0.9344} & \textbf{45.18}  \\
    \bottomrule
    \end{tabular}
}
\end{minipage}\hfill
\vspace{-1em}
\end{figure*}

\subsubsection{Real-world Data Generation}
We build our fitting pipeline for data generation to enable arbitrary novel views rendering from multi-camera captured data, which 
serve as supervision to train our framework.
This pipeline is built upon the 3D Gaussian Splatting (3DGS) representation~\cite{kerbl20233d}, which is particularly suitable for fitting because of its fast optimization and real-time rendering capability.
Specifically, 3DGS represents a scene as a set of 3D Gaussians, each defined by a center position $p$ and a covariance matrix $\Sigma$, capturing  
spatial location and shape:
\begin{equation}
    G(x) = \exp\left(-\frac{1}{2}(x - p)^{T} \Sigma^{-1}(x - p)\right),
\end{equation}
where $x$ is a point in 3D space. These Gaussians are projected onto the image plane and composited using a tile-based rasterizer with alpha blending, supporting differentiable and real-time rendering.

Given a set of images $\mathcal{I} = \{ I_i \mid i \in [1, M] \}$ captured from multiple-view datasets (\eg DNA-Rendering~\cite{cheng2023dna} and MVHumanNet~\cite{xiong2024mvhumannet}) and the corresponding SMPL-X mesh of the human subject, we initialize the 3D Gaussians by assigning one Gaussian to each mesh vertex. The center of each Gaussian is set to the vertex position, and its covariance is initialized accordingly.
We then optimize the parameters of these Gaussians by minimizing the photometric reconstruction loss:
\begin{equation}
\mathcal{L} = \left\| I_i - f\big(V(\theta), \pi_i\big) \right\|^2,
\end{equation}
where $V(\theta)$, $\pi_i$, and $f(\cdot)$ denote the set of Gaussian parameters, the camera parameters and the differentiable rendering function, respectively. 
During training, we sample one view per iteration and apply adaptive density control~\cite{kerbl20233d} to improve convergence and coverage.

After fitting, we obtain a refined 3D Gaussian representation of the posed human, which we denote with $G_p$.  
Finally, we render multiple images from the re-centered $G_p$ under a predefined set of canonical viewpoints, which could serve as a data source for training.

%% file: sec/4_experiments.tex
\section{Experiments}
\label{sec_experiment}
In this section, we present our experimental evaluation. We begin by describing our experimental setup, which includes the datasets, implementation details, and evaluation metrics. 
Then, we compare our method with state-of-the-art approaches both quantitatively and qualitatively.
Finally, we conduct ablation studies for the most important components of our approach.

\subsection{Experimental Setup}
\label{sec_setup}

\noindent \textbf{Implementation details.}
Our framework is implemented on PyTorch.
The experiments are conducted on NVIDIA H100, and we utilize 64 H100s for training.
The hyperparameters $\lambda_m$ and $\lambda_p$ are both set to 0.5.
During training, we set $N=4$ as the number of rendered views.
We utilize AdamW as the optimizer.
The learning rate is set as 6e-4 with the batch size as 64.
The triplane spatial size in this work is 96.
The cropped patch size during training is 180.

\noindent \textbf{Datasets.}
During training, we utilize two kinds of data sources, \emph{i.e.,} commonly utilized human scans (THuman2~\cite{yu2021function4d}, CustomHuman~\cite{shen2023xavatar}, and 2K2K~\cite{han2023high}) and our generated data.
The former size contains around 5k assets with 180k images, while the latter one has 100k assets with 2.6M images.
Evaluation is performed on three benchmarks, including THuman2, CustomHuman, and 2K2K datasets.

\noindent \textbf{Evaluation metrics.}
Following previous works~\cite{pan2024humansplat,albahar2023single,he2025magicman}, we evaluate the reconstructed avatar on rendering quality.
During evaluation of rendering quality, we select three widely-used metrics, \emph{i.e.,} Peak Signal-to-Noise Ratio (PSNR), Structural Similarity Index Measure (SSIM)~\cite{wang2004image}, and Learned Perceptual Similarity (LPIPS)~\cite{zhang2018unreasonable}.
For a fair comparison, the resolution of rendered images is set to $512\times512$.
Our rendered views are uniformly distributed around the human subject using 20-degree intervals.
Besides the most common evaluation via view synthesis metrics, we additionally report Chamfer Distance (CD), Normal Consistency (NC), and F-Score as secondary metrics.
These metrics offer complementary insights into the geometric quality of the reconstructed 3D results.

\begin{table}[t]
    \footnotesize
    \centering
    \setlength{\tabcolsep}{2pt}
    \caption{{Effectiveness of our generated data}. It is validated via fine-tuning another method (Real3D~\cite{jiang2024real3d}) with our generated data.}
    \resizebox{1.0\linewidth}{!}{
    \begin{tabular}{l ccc ccc ccc}
    \toprule
    \multirow{2}{*}{\textbf{Method}}  & \multicolumn{3}{c}{\textbf{CustomHuman}\quad\quad} & \multicolumn{3}{c}{\textbf{THuman2}\quad\quad} & \multicolumn{3}{c}{\textbf{2K2K}\quad\quad} \\ 
    \cmidrule(lr){2-4} \cmidrule(lr){5-7} \cmidrule(lr){8-10}
    & {\bf PSNR}$\uparrow$ & {\bf SSIM}$\uparrow$ & {\bf LPIPS}$\downarrow$ & {\bf PSNR}$\uparrow$ & {\bf SSIM}$\uparrow$ & {\bf LPIPS}$\downarrow$ & {\bf PSNR}$\uparrow$ & {\bf SSIM}$\uparrow$ & {\bf LPIPS}$\downarrow$ \\ \midrule
    \multicolumn{3}{l}{\emph{Input: Frontal view}} \\
    {Real3D}  & 17.13 & 0.8990 & 95.12 & 19.14 & 0.9094 & 87.68 & 18.06 & 0.9020 & 81.78 \\
    {Real3D (+ our data)}  & \textbf{20.97} & \textbf{0.9268} & \textbf{58.54}	& \textbf{23.10} & \textbf{0.9325} & \textbf{55.30} & \textbf{20.91} & \textbf{0.9202} & \textbf{58.22} \\   
    \multicolumn{3}{l}{\emph{Input: Side view}} \\
    {Real3D}  & 17.13 & 0.8990 & 95.12 & 19.14 & 0.9094 & 87.68 & 18.06 & 0.9020 & 81.78 \\
    {Real3D (+ our data)}  & \textbf{19.46} & \textbf{0.9113} & \textbf{66.09} & \textbf{22.28} & \textbf{0.9287} & \textbf{57.20} & \textbf{20.47} & \textbf{0.9142} & \textbf{58.14} \\  
    \bottomrule
    \end{tabular}
    }
    \label{tab:ablation_data}
    \vspace{-1.2em}
\end{table}

\begin{table*}[t!]
    \footnotesize
    \centering
    \setlength{\tabcolsep}{12pt}
    \caption{\textbf{Comparison with previous state-of-the-art methods on rendering quality.} These include Real3D, SF3D, Trellis, Hunyuan2, PaMIR, SiFU and SiTH, on the CustomHuman, THuman2 and 2K2K datasets. We outperform all previous methods across all evaluated metrics with a notable gain. $\uparrow$ and $\downarrow$ represent the higher the better, and the lower the better, respectively.}
    \vspace{-0.5em}
    \resizebox{0.98\linewidth}{!}{
    \begin{tabular}{l ccc ccc ccc}
    \toprule
    \multirow{2}{*}{\textbf{Method}}  & \multicolumn{3}{c}{\textbf{CustomHuman}\quad\quad} & \multicolumn{3}{c}{\textbf{THuman2}\quad\quad} & \multicolumn{3}{c}{\textbf{2K2K}\quad\quad} \\ 
    \cmidrule(lr){2-4} \cmidrule(lr){5-7} \cmidrule(lr){8-10}
    & {\bf PSNR}$\uparrow$ & {\bf SSIM}$\uparrow$ & {\bf LPIPS}$\downarrow$ & {\bf PSNR}$\uparrow$ & {\bf SSIM}$\uparrow$ & {\bf LPIPS}$\downarrow$ & {\bf PSNR}$\uparrow$ & {\bf SSIM}$\uparrow$ & {\bf LPIPS}$\downarrow$ \\ \midrule
    \multicolumn{3}{l}{\emph{Input: Frontal view}} \\
    {Real3D~\cite{jiang2024real3d}}  & 17.13 & 0.8990 & 95.12 & 19.14 & 0.9094 & 87.68 & 18.06 & 0.9020 & 81.78 \\
    {SF3D~\cite{boss2024sf3d}}  & 19.46 & 0.9113 & 66.09 & 22.28 & 0.9287 & 57.20 & 20.47 & 0.9142 & 58.14 \\
    {Trellis~\cite{xiang2024structured}}  & 18.59 & 0.9123 & 74.98 & 20.77 & 0.9218 & 65.67 & 19.21 & 0.9140 & 68.25 \\
    {Hunyuan2~\cite{zhao2025hunyuan3d}}  & 19.42 & 0.9094 & 74.34 & 21.44 & 0.9257 & 66.19 & 19.87 & 0.9145 & 65.62 \\
    {PaMIR~\cite{zheng2021pamir}} & 18.15 & 0.9070 & 88.12 & 21.03 & 0.9229 & 70.91 & 18.89 & 0.9113 & 73.90 \\
    {SiFU~\cite{zhang2024sifu}}  & 17.94 & 0.9091 & 85.75 & 19.44 & 0.9157 & 79.62 & 16.82 & 0.9039 & 87.51\\
    {SiTH~\cite{ho2024sith}} & 19.13 & 0.9173 & 72.94 & 20.92 & 0.9231 & 66.90 & 18.49 & 0.9095 & 73.55 \\
    {\NAME} & {\bf 22.29} & {\bf 0.9360} & {\bf 42.42} & {\bf 23.96} & {\bf 0.9382} & {\bf 42.13} & {\bf 22.65} & {\bf 0.9336} & {\bf 41.72}  \\
    \midrule
    \multicolumn{3}{l}{\emph{Input: Side view}} \\
    {Real3D~\cite{jiang2024real3d}} & 17.42 & 0.9005 & 94.94 & 19.40 & 0.9109 & 87.90 & 18.67 & 0.9047 & 78.54 \\
    {SF3D~\cite{boss2024sf3d}}  & 19.05 & 0.9085 & 69.75 & 21.25 & 0.9227 & 65.02 & 20.50 & 0.9139 & 57.85 \\
    {Trellis~\cite{xiang2024structured}}  & 17.21 & 0.9071 & 88.08 & 19.50 & 0.9172 & 74.79 & 18.12 & 0.9088 & 74.57 \\
    {Hunyuan2~\cite{zhao2025hunyuan3d}}  & 19.22 & 0.9076 & 77.28 & 21.01 & 0.9234 & 70.02 & 19.81 & 0.9139 & 66.48 \\
    {PaMIR~\cite{zheng2021pamir}} & 17.24 & 0.9006 & 97.10 & 20.59 & 0.9190 & 76.41 & 18.27 & 0.9077 & 78.95 \\
    {SiFU~\cite{zhang2024sifu}}  & 17.25 & 0.9040 & 93.26 & 19.32 & 0.9145 & 82.28 & 17.24  & 0.9052 & 85.49  \\
    {SiTH~\cite{ho2024sith}}  & 18.25 & 0.9114 & 81.95 & 20.43 & 0.9201 & 71.92 & 18.72 & 0.9102 & 73.25 \\
    {\NAME} & {\bf 22.52} & {\bf 0.9348} & {\bf 43.58} & {\bf 24.35} & {\bf 0.9382} & {\bf 42.55} & {\bf 23.07} & {\bf 0.9347} & {\bf 40.95} \\   
    \bottomrule
    \end{tabular}
    }
    \label{tab:comparison}
    \vspace{-1em}
\end{table*}

\subsection{Ablation Studies}
We perform ablation studies to highlight the most important components in our proposed method, including data generation and model settings.
Unless stated otherwise, training runs for 70\% of the total iterations adopted in the main comparison.
The evaluation is conducted on the CustomHuman dataset with the frontal view as the input.

\noindent \textbf{Effectiveness of our generated data.}
Our generated data includes two complementary components: \emph{gen-data (assets)}, i.e., data generated from rigged assets, and \emph{gen-data (multi-cam)}, which is generated from fitting real-world multi-camera data.
As shown in Table~\ref{tab:ablation1}, both components contribute effectively to the training process, enhancing the robustness and diversity of the model. 
Notably, synthetic assets can provide extensive diversity on the pose, given the animation strategy.
Besides, the real-world data, though smaller in quantity, provides important signals for achieving high-quality reconstruction, especially in terms of photorealism. 
We believe this is due to the natural appearance statistics and complex visual cues present in real-world images, which are difficult to fully simulate through synthetic data alone.
To further analyze the impact of data scale, we vary the portion of generated data used during training. As shown in Table~\ref{tab:ablation2}, model performance improves consistently with more generated data. 

We further provide evidence that could validate the effectiveness of our generated data via fine-tuning Real3D~\cite{jiang2024real3d} (a general 3D object reconstruction method) on our proposed dataset.
As shown in Table~\ref{tab:ablation_data}, it can be observed that Real3D could benefit from our proposed new data with a notable performance gain on multiple benchmarks.

\noindent \textbf{Impact of model settings.}
We begin by investigating the impact of incorporating a mesh prior into our pipeline (Table~\ref{tab:ablation3}). 
When integrated, the mesh prior yields a 2.3\% relative improvement in LPIPS, demonstrating that, even with a large-scale training dataset, mesh provides a valuable cue for capturing detailed human structural features. 
In addition, we adopt the triplane representation as the final 3D depiction, which compresses 3D geometry by decomposing the volume into three orthogonal projection planes. 
As shown in Table~\ref{tab:ablation3}, reducing the triplane spatial size to 32, a common setting in LRMs, results in a significant performance drop. 
This finding highlights the necessity for a model capacity that aligns with the demanding requirements of human reconstruction, where capturing fine-grained details is more critical compared to general object reconstruction.

\subsection{Comparison with State-of-the-art Methods}
\textbf{Baselines.} The baselines we use in our evaluation include methods for both general object reconstruction and human-specific reconstruction from a single image.
Specifically, Real3D~\cite{jiang2024real3d} and SF3D~\cite{boss2024sf3d} are LRM-based state-of-the-art methods in general object modeling, which can achieve fast reconstruction with a single feed-forward inference.
PaMIR~\cite{zheng2021pamir}, SiFU~\cite{zhang2024sifu} and SiTH~\cite{ho2024sith} require instance-level post-processing to return the 3D human, 
usually involving techniques like diffusion, mesh fitting, \emph{etc.}
However, these stages lead to a significant increase in the inference time.
Trellis~\cite{xiang2024structured} and Hunyuan2~\cite{zhao2025hunyuan3d} are two very recent large-scale industrial foundation models that perform flow-based generative modeling in 3D space, trained on around 1 million diverse 3D assets to achieve high-quality generation.

\begin{figure*}[t!]
    \centering
    \includegraphics[width=\linewidth]{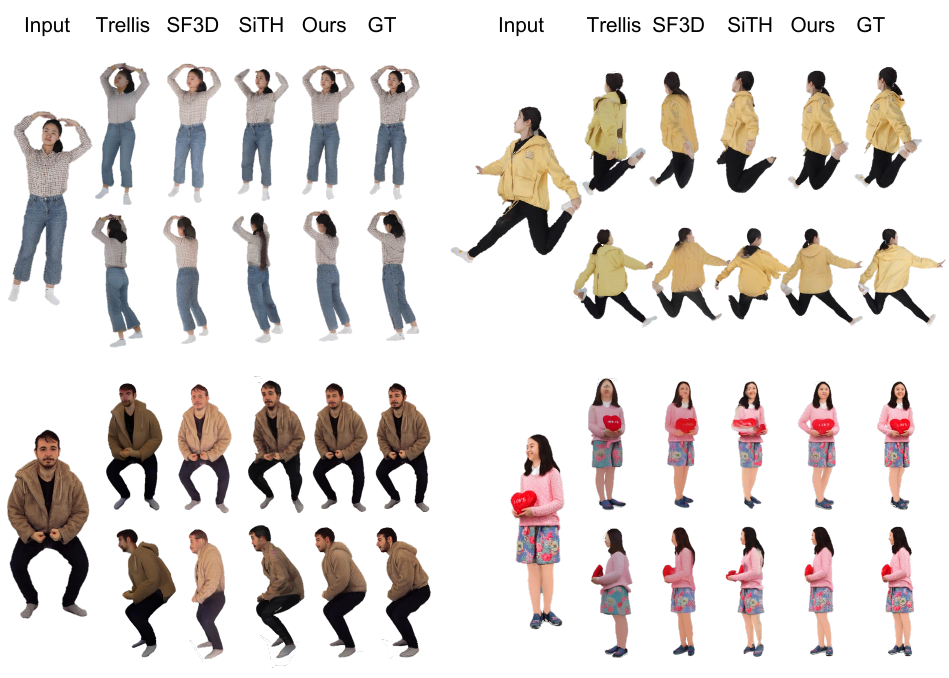}
    \includegraphics[width=\linewidth]{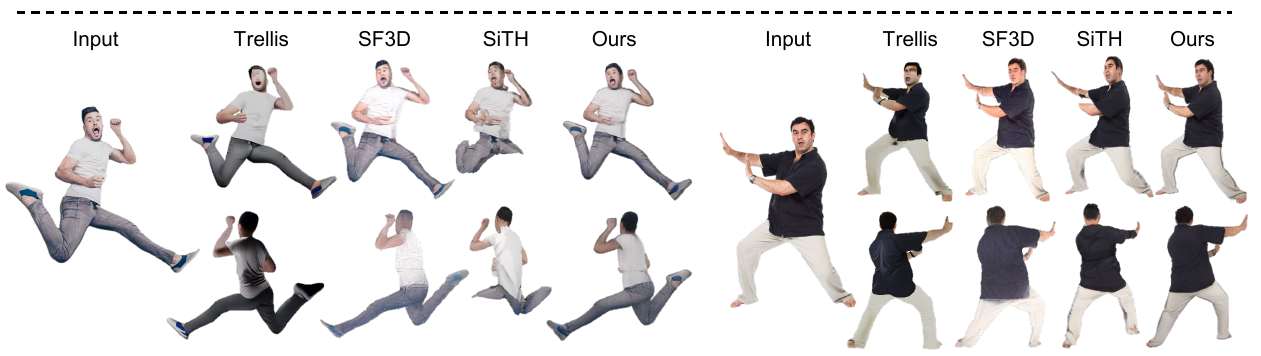}
    \vspace{-2em}
    \caption{\textbf{Qualitative comparison with state-of-the-art methods}, including input from \emph{benchmarks} (top) and \emph{in-the-wild images} (bottom). The reconstructed human by our \NAME{} method shows superior structure and appearance. (Best viewed in color.)}
    \label{fig:compare}
    \vspace{-1em}
\end{figure*}

\noindent \textbf{Evaluation Protocol.} Fair comparison is considered via following \emph{established practice in this field}: each method is trained using the settings from its original paper, and all methods are then evaluated on the same benchmark dataset for comparison.
This is consistent with prior works (\eg, SiTH, SiFU, Trellis, Hunyuan2) and allows a fair comparison of generalization ability and reconstruction quality.

The experiments are conducted on multiple benchmarks \emph{i.e.,} CustomHuman~\cite{shen2023xavatar}, THuman2~\cite{yu2021function4d} and 2K2K~\cite{han2023high}.
Existing benchmarks are built based on normalized human scans, such that the canonical view is defined as the subject's frontal perspective. 
However, in real-world scenarios, it is not guaranteed that the input images will always show a frontal view.
To address this discrepancy, we additionally incorporate side-view data into our experimental evaluations, which helps assess the robustness and generalizability of our approach.
Specifically, we render images from a viewpoint that has a horizontal offset of 60 degrees relative to the canonical view and use them as input.
Considering that some methods (SiTH, SiFU, Trellis, and Hunyuan2) output mesh as the final representations, we adopt a fair comparison protocol by first aligning the scale of the reconstructed meshes to that of the ground truth. 
Subsequently, we apply an Iterative Closest Point (ICP) alignment to further refine the correspondence between the reconstructions and the ground truth.
Finally, we utilize the re-aligned mesh for rendering and geometry evaluation.

\noindent \textbf{Results.} As shown in Table~\ref{tab:comparison}, we observe that our method outperforms all previous approaches under different input conditions.
Compared with the best competitor SiTH, our method achieves 41.8\%, 37.0\% and 43.3\% relative LPIPS gain on the CustomHuman, THuman2 and 2K2K datasets, respectively.
It is evident that previous general object reconstruction methods like Real3D, SF3D and Hunyuan2 are moderately robust to variations in viewpoint.
However, they suffer from suboptimal performance due to the lack of specialized training on human-specific data. 
Additionally, previous state-of-the-art human reconstruction approaches often yield lower quality results when given a side-view input, as they do not generalize well to such less common viewpoints.
In contrast, our approach yields superior performance in both appearance and structure.

\begin{table*}[t!]
    \footnotesize
    \centering
    \setlength{\tabcolsep}{12pt}
    \caption{\textbf{Additional evaluation on geometry quality.} When reporting CD, we consider both the prediction to ground truth and the ground truth to prediction distance. We outperform all previous methods across all evaluated metrics with a notable gain.  $\uparrow$ and $\downarrow$ represent that higher is better, and that lower is better, respectively.}
    \vspace{-0.5em}
    \resizebox{1.0\linewidth}{!}{
    \begin{tabular}{l ccc ccc ccc}
    \toprule
    \multirow{2}{*}{\textbf{Method}}  & \multicolumn{3}{c}{\textbf{CustomHuman}\quad\quad} & \multicolumn{3}{c}{\textbf{THuman2}\quad\quad} & \multicolumn{3}{c}{\textbf{2K2K}\quad\quad} \\ 
    \cmidrule(lr){2-4} \cmidrule(lr){5-7} \cmidrule(lr){8-10}
    & {\bf CD}$\downarrow$ & {\bf NC}$\uparrow$ & {\bf F-Score}$\uparrow$ & {\bf CD}$\downarrow$ & {\bf NC}$\uparrow$ & {\bf F-Score}$\uparrow$ & {\bf CD}$\downarrow$ & {\bf NC}$\uparrow$ & {\bf F-Score}$\uparrow$ \\ \midrule
    \multicolumn{3}{l}{\emph{Input: Frontal view}} \\
    {SF3D~\cite{boss2024sf3d}}  & 1.738/2.040 & 0.847 & 39.585 & 1.441/1.745 & 0.833 & 43.820 & 1.204/1.412 & 0.829 & 50.900 \\
    {Trellis~\cite{xiang2024structured}}  & 2.125/2.175 & 0.801 & 32.846 & 1.799/1.832 & 0.796 & 37.939 & 1.446/1.359 & 0.805 & 48.826 \\
    {Hunyuan2~\cite{zhao2025hunyuan3d}}  & 1.799/1.762 & 0.837 & 38.365 & 1.562/1.541 & 0.808 & 43.868 & 1.237/1.217 & 0.829 & 53.946 \\
    {ICON~\cite{xiu2022icon}}  & 2.468/2.915 & 0.779 & 27.731 & 2.568/3.168 & 0.752 & 26.453 & 2.211/3.331 & 0.728 & 28.805 \\
    {ECON~\cite{xiu2023econ}}  & 2.160/2.813 & 0.804 & 33.429 & 2.240/3.931 & 0.763 & 31.294 & 2.066/6.232 & 0.732 & 32.927 \\
    {SiFU~\cite{zhang2024sifu}}  & 2.440/3.203 & 0.789 & 27.553 & 2.509/3.778 & 0.760 & 27.487 & 2.136/5.331 & 0.732 & 29.823 \\
    {SiTH~\cite{ho2024sith}} & 1.792/2.215 & 0.826 & 36.822 & 1.741/2.082 & 0.805 & 39.666 & 1.518/1.896 & 0.798 & 42.859 \\
    {\NAME} & {\bf 1.062/1.102} & {\bf 0.867} & {\bf 61.379} & {\bf 1.027/1.098} & {\bf 0.840} & {\bf 61.939} & {\bf 1.045/1.110} & {\bf 0.836} & {\bf 60.673}  \\
    \midrule
    \multicolumn{3}{l}{\emph{Input: Side view}} \\
    {SF3D~\cite{boss2024sf3d}}  & 1.657/2.177 & 0.834 & 39.931 & 1.695/2.284 & 0.810 & 39.102 & 1.234/1.452 & 0.823 & 49.584 \\
    {Trellis~\cite{xiang2024structured}} & 2.361/4.802 & 0.757 & 29.603 & 2.089/2.933 & 0.761 & 34.754 & 1.606/1.644 & 0.772 & 44.555 \\
    {Hunyuan2~\cite{zhao2025hunyuan3d}} & 1.848/1.898 & 0.827 & 37.698 & 1.642/1.677 & 0.811 & 42.618 & 1.259/1.235 & 0.821 & 54.622 \\
    {ICON~\cite{xiu2022icon}}  & 2.536/3.635 & 0.769 & 28.138 & 2.286/3.012 & 0.760 & 30.981 & 1.892/2.873 & 0.745 & 35.828 \\
    {ECON~\cite{xiu2023econ}}  & 2.592/5.233 & 0.756 & 27.284 & 2.270/4.747 & 0.750 & 31.298 &  1.811/5.778 & 0.739 & 36.933 \\
    {SiFU~\cite{zhang2024sifu}}  & 2.615/6.112 & 0.756 & 24.788 & 2.301/4.631 & 0.748 & 30.057 & 1.820/5.550 & 0.736 & 35.653  \\
    {SiTH~\cite{ho2024sith}}  & 2.063/3.219 & 0.793 & 32.189 & 1.849/2.557 & 0.789 & 36.070 & 1.633/1.879 & 0.791 & 40.807 \\
    {\NAME} & {\bf 1.052/1.079} & {\bf 0.858} & {\bf 62.553} & {\bf 1.044/1.123} & {\bf 0.829} & {\bf 62.344} & {\bf 1.058/1.133} & {\bf 0.829} & {\bf 60.126} \\   
    \bottomrule
    \end{tabular}
    }
    \label{tab:comparison_geo}
    \vspace{-1.0em}
\end{table*}

\begin{figure}[t]
    \centering
    \includegraphics[width=\linewidth]{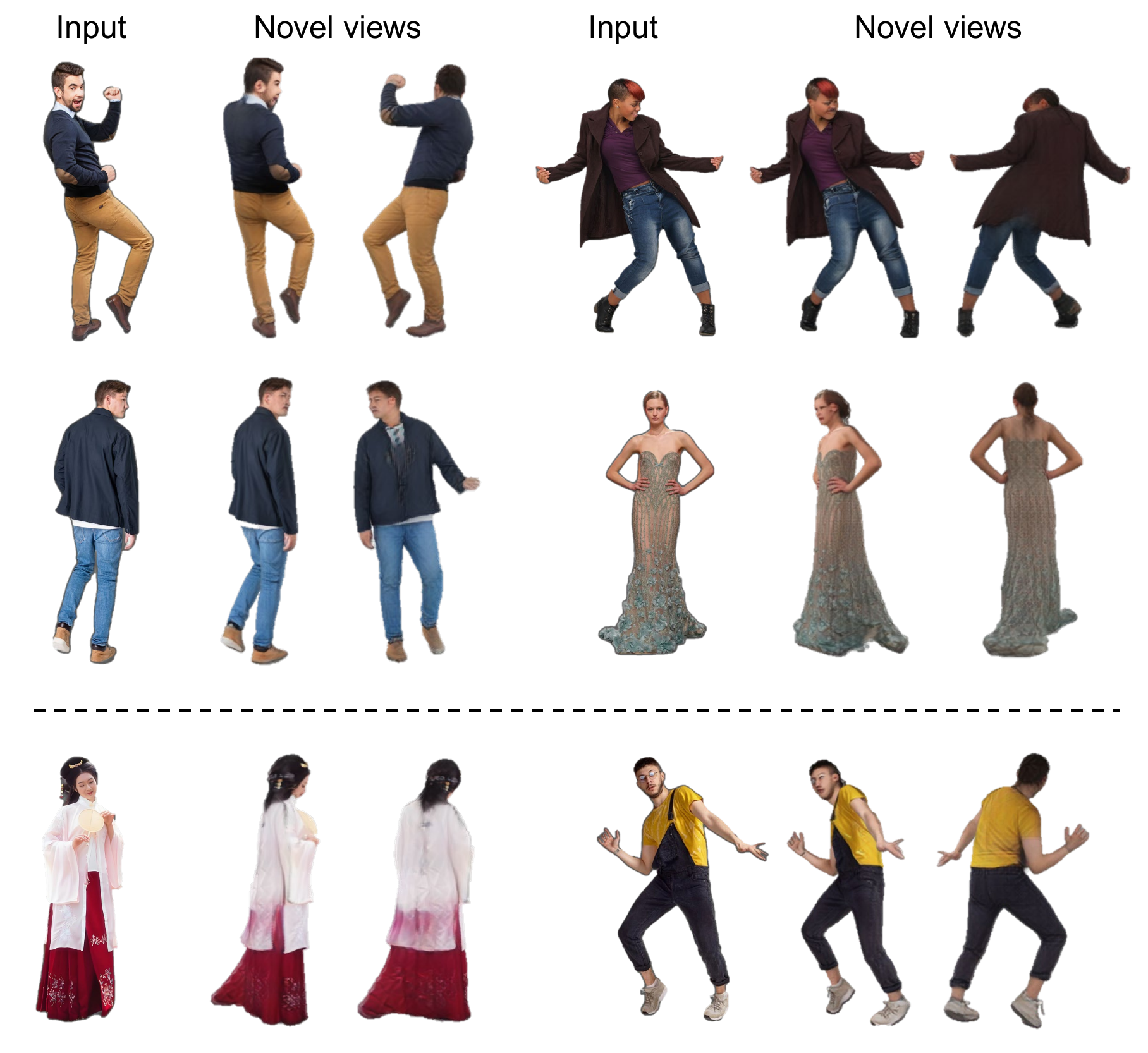}
    \vspace{-1em}
    \caption{\textbf{Qualitative evaluation of our approach with in-the-wild images as input.} We also show some typical failure cases (bottom), \emph{e.g.,} inferring the plausible back texture of challenging clothes like dresses and overalls. (Best viewed in color.)}
    \label{fig:wild}
    \vspace{-1em}
\end{figure}

The qualitative comparison in Figure~\ref{fig:compare} also validates the effectiveness of our method.
Beyond the results presented in Figure~\ref{fig:teaser}, we also showcase results on a diverse set of in-the-wild images in Figure~\ref{fig:wild}. 
These images include a wide range of poses and clothing styles, demonstrating that our method effectively captures fine details and accurately represents the 3D structure even in challenging conditions. 
Notably, our method could even reconstruct a person interacting with an object.
However, we also observe certain failure cases; for example, the network sometimes struggles to accurately infer plausible textures on the back of the subjects. Addressing these shortcomings may require the adoption of more powerful models in future work.

As shown in Table~\ref{tab:comparison_geo}, \NAME{} consistently achieves the best performance across all metrics.
Notably, on the side-view inputs, where occlusion affects performance, \NAME{} outperforms the human-specific method SiTH by 94.3\% relative F-Score gain on CustomHuman.

Due to space limitations, we provide additional discussions and results in the \emph{Supplementary Material}. These include: (1) further analysis of the quality of our generated data, (2) comparisons with animation-based approaches, (3) more ablations and (4) more details of our experimental setup, such as data generation, \NAME{} training, and patch selection.
We also include extended qualitative results. More visual results and videos can be found in the accompanying supplementary video.

%% file: sec/5_conclusion.tex
\section{Conclusion}
\label{sec_conclusion}
In this work, we present \NAME{}, a large human avatar reconstruction model that enables universal and photorealistic 3D single-view human reconstruction in less than a second.
Our contributions are two-fold, focusing on data generation and model design.
To mitigate the scarcity of 3D human data, we design an effective and scalable data generation pipeline that integrates multiple sources (e.g., 3D assets and multi-camera captures), expanding the training dataset to 100k assets and significantly improving its diversity and scale.
Built on this large-scale data, \NAME~leverages the estimated simplified human mesh as an extra prior and directly lifts the input single image to a triplane representation that effectively captures the details of the 3D human.
Extensive experiments on three benchmark datasets show that our approach consistently outperforms existing methods, with over 40\% improvement in LPIPS and more than 90\% relative gain in F-Score.
Future work could focus on addressing the challenges of complex real-world scenarios, \eg, dealing with occlusion in the input image, extending reconstruction to human interactions, \emph{etc}.

\medskip
\noindent 
\textbf{Acknowledgments.}
This research has been supported by the Sony Research Award Program and computing support on the Vista GPU Cluster through the Center for Generative AI (CGAI) and the Texas Advanced Computing Center (TACC) at 
UT
Austin.
GP was supported by NSF IIS-2504906, IIS-2544200 and Gifts from Google and Adobe.

%% file: sec/6_supp.tex
In this Supplementary Material, we provide additional details that were not included in the main manuscript due to space constraints. Specifically, we include more discussions (Section~\ref{sec:discuss}), more details on the experiment setup (Section~\ref{sec:setup_supp}), and more visual results (Section~\ref{sec:visual}).

\section*{A. More Discussions}
\label{sec:discuss}

\noindent \textbf{Quality of our generated data.}
Our generated dataset contains two components, \emph{i.e.,} the synthetic subset and the real-world subset.
For the synthetic subset, we render synthetic multi-view images in Blender from rigged SynBody~\cite{yang2023synbody} assets. Because SynBody meshes and textures are of high quality and Blender’s renderer preserves their visual fidelity, the resulting images maintain the same level of detail and realism as the source assets.
For the real-world subset, this subset is generated by fitting multi-view capture data to 3DGS. We then re-render the fitted 3DGS back to the capture viewpoints and compare them with the corresponding ground-truth images, achieving an average of 36.23 / 0.9881 / 16.57 (PSNR / SSIM / LPIPS).

\noindent \textbf{Design of real-world data generation.}
We argue that good initialization is critical in the real-world data generation stage. 
COLMAP initialization is the common practice in 3DGS, but it performs poorly on multi-view human capture data. 
To illustrate this, we performed a test on 10 randomly selected samples. 
Our improved initialization based on SMPL-X vertices enables significantly fewer training iterations, reducing optimization time from 40 minutes to 4 minutes.
The performance on the initialization method is shown in Table~\ref{tab:init}.

\begin{table}[h]
    \footnotesize
    \centering
    \setlength{\tabcolsep}{12pt}
    \caption{\textbf{Impact of initialization in real-world data generation.} $\uparrow$ and $\downarrow$ represent the higher the better, and the lower the better, respectively.}
    \begin{tabular}{l ccc}
    \toprule
    \textbf{Method} & {\bf PSNR}$\uparrow$ & {\bf SSIM}$\uparrow$ & {\bf LPIPS}$\downarrow$ \\ \midrule
    {COLMAP} & 16.49 & 0.9025 & 68.79 \\
    {Ours} & \textbf{36.38} & \textbf{0.9886} & \textbf{16.36} \\
    \bottomrule
    \end{tabular}
    \label{tab:init}
\end{table}

\noindent \textbf{Comparison with animation-based methods.}
Methods like SHERF~\cite{hu2023sherf} and LHM~\cite{qiu2025lhm} fall under a related but distinct task setting: they aim to build generalizable, animatable human avatars from a single image. Since these methods focus primarily on pose-driven animation, they typically rely on ground-truth, high-quality SMPL/SMPL-X poses during both training and testing to ensure precise alignment.

In contrast, our task operates under a more challenging input condition: only a single image is available. That is to say, SMPL(-X) parameters must be estimated using off-the-shelf tools. 
For SHERF and LHM, the SMPL(-X) pose is derived from 4D Humans (same as our method) and its provided demo, respectively. We follow SHERF's instructions from their Github repo and train it on the corresponding dataset. 
For LHM, since it does not release training and testing code, we directly utilize its pre-trained weights and its demo code to perform inference.
As shown in Table~\ref{tab:anim}, 
our method achieves superior performance with notable gains across multiple datasets and input viewpoints. For LHM and SHERF, they need to animate the avatar from the canonical space to the target pose space, during which any pose misalignment will directly exist in the rendering results, leading to performance degradation.

\noindent \textbf{Generalization.}
For evaluation, we use a single model and report its performance across all benchmarks.
To further assess the generalization capability of HumanNOVA, we conduct a leave-one-out evaluation on the CustomHuman dataset.
As shown in Table~\ref{tab:generalization}, the performance under the leave-one-out setting remains comparable to that of the full model, indicating that HumanNOVA generalizes well to unseen settings.

\begin{table}[t]
    \footnotesize
    \centering
    \setlength{\tabcolsep}{12pt}
    \caption{\textbf{Comparison with animation-based methods.} $\uparrow$ and $\downarrow$ represent the higher the better, and the lower the better, respectively.}
    \begin{tabular}{l ccc}
    \toprule
    \textbf{Method} & {\bf PSNR}$\uparrow$ & {\bf SSIM}$\uparrow$ & {\bf LPIPS}$\downarrow$ \\ \midrule
    {SHERF~\cite{hu2023sherf}} & 16.83 & 0.9037 & 87.99 \\
    {LHM~\cite{qiu2025lhm}} & 17.75 & 0.9083 & 76.85 \\
    {Ours} & \textbf{22.29} & \textbf{0.9360} & \textbf{42.42} \\
    \bottomrule
    \end{tabular}
    \label{tab:anim}
\end{table}

\begin{figure}[t]
    \centering
    \includegraphics[width=\linewidth]{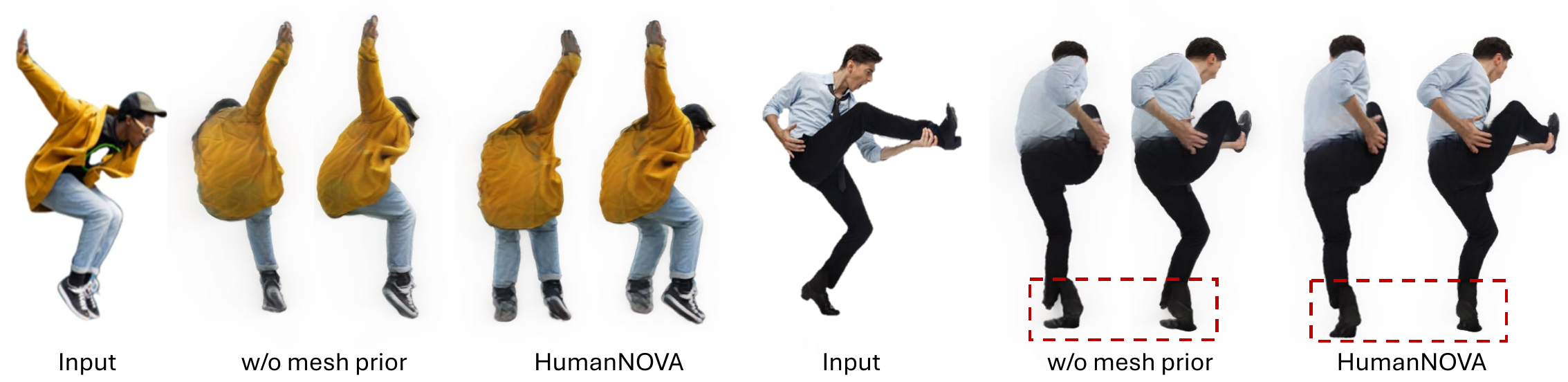}
    \vspace{-1em}
    \caption{Visual results on the effectiveness of the SMPL prior.}
    \label{fig:smpl}
\end{figure}

\begin{figure}[t]
    \centering
    \includegraphics[width=\linewidth]{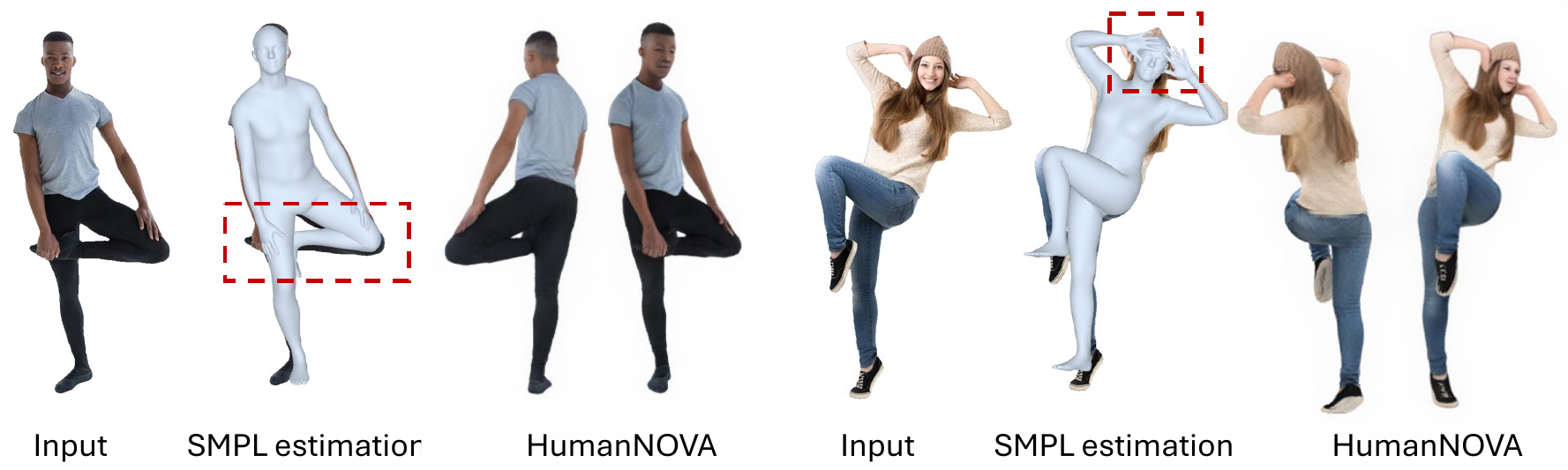}
    \vspace{-1em}
    \caption{Visual results on the robustness of HumanNOVA under inaccurate SMPL estimates.}
    \label{fig:robust}
\end{figure}

\noindent \textbf{More analysis on SMPL mesh prior.}
We provide more visualization about the effect of the mesh prior in Figure~\ref{fig:smpl}.
It can be observed that the base LRM architecture (w/o mesh prior) fails to predict feasible human structures, while HumanNOVA performs robustly when the input is an in-the-wild image.
In Figure~\ref{fig:robust}, we demonstrate two representative in-the-wild examples (backgrounds are removed) where the estimated SMPL is visibly inaccurate. It can be observed that HumanNOVA remains resilient to this noise and adaptively prioritizes the appearance cues. 
However, our approach can still fail in the extreme case where the SMPL estimation completely breaks.

\noindent \textbf{More ablations on HumanNOVA.}
We conduct additional ablation studies on the CustomHuman dataset to analyze the impact of different design choices in HumanNOVA.
As shown in Table~\ref{tab:ablation_more}, utilizing Sapiens~\cite{khirodkar2024sapiens} as the visual encoder still achieves comparable performance with DINOv2.
Reducing the number of fusion layers from 4 to 2 causes the most significant performance drop, especially in LPIPS, suggesting that sufficient cross-modal fusion depth is critical for human avatar modeling.
In addition, decreasing the number of supervision views from 4 to 2 also results in inferior performance, which demonstrates the importance of richer multi-view supervision for improving reconstruction quality.

\begin{table}[t]
    \footnotesize
    \centering
    \setlength{\tabcolsep}{8pt}
    \caption{\textbf{Generalization capability of HumanNOVA.} We conduct a leave-one-out evaluation with the CustomHuman dataset. $\uparrow$ and $\downarrow$ represent the higher the better, and the lower the better, respectively.}
    \begin{tabular}{l ccc}
    \toprule
    \textbf{Method} & {\bf PSNR}$\uparrow$ & {\bf SSIM}$\uparrow$ & {\bf LPIPS}$\downarrow$ \\ \midrule
    {Leave-CustomHuman-Out} & 21.99 & 0.9344 & 44.22 \\
    {Ours} & \textbf{22.29} & \textbf{0.9360} & \textbf{42.42} \\
    \bottomrule
    \end{tabular}
    \label{tab:generalization}
\end{table}

\begin{table}[t!]
    \footnotesize
    \centering
    \setlength{\tabcolsep}{4pt}
    \caption{\textbf{More ablations on HumanNOVA framework on the CustomHuman dataset.} $\uparrow$ and $\downarrow$ represent the higher the better, and the lower the better, respectively.}
    \begin{tabular}{l ccc}
    \toprule
    \textbf{Method} & {\bf PSNR}$\uparrow$ & {\bf SSIM}$\uparrow$ & {\bf LPIPS}$\downarrow$ \\ \midrule
    {Ours} & \textbf{22.29} & \textbf{0.9360} & \textbf{42.42} \\
    {Visual encoder (DINOv2 -> Sapiens)} & 21.98 & 0.9327 & 46.52 \\
    {Fusion layer (4 -> 2)} & 21.42 & 0.9301 & 50.65 \\
    {Supervision views (4 -> 2)} & 22.07 & 0.9344 & 45.18 \\
    \bottomrule
    \end{tabular}
    \label{tab:ablation_more}
\end{table}

\noindent \textbf{Broader Impact.}
We are dedicated to fast and photorealistic 3D human avatar modeling from a single image.
However, these technological advancements also raise concerns regarding potential misuse, including deceptive practices, harassment, and privacy violations. 
The lowered barrier to generating realistic human reconstructions could facilitate the creation of harmful or unauthorized content, thereby intensifying privacy risks. 
This highlights the need for careful consideration of both technical and regulatory measures.

\section*{B. More Experiment Setup Details}
\label{sec:setup_supp}

\noindent \textbf{Our data generation.} For synthetic data generation, we utilize all 1,000 unique characters officially released by SynBody~\cite{yang2023synbody} for public download. These assets exhibit high diversity in terms of ethnicity, body shape, and clothing. Specifically, the characters span a wide range of skin tones, and wear diverse outfits based on approximately 68 clothing templates, including dresses, T-shirts, coats, pants, and more.
For real-world data generation, the Gaussian optimization only lasts 4,000 iterations with the densification performed between iterations 400 and 1,500.
For other parameters, we follow the original settings~\cite{kerbl20233d}.
Note that both data generation strategies are scalable.
The data size of the generated real-world data is 22k assets, while the generated synthetic data contains 78k assets.
We generate an average of 26 views per asset, with camera positions randomly distributed on a sphere. The azimuth angles range from $0^\circ$ to $360^\circ$, and the elevation angles range from $-45^\circ$ to $60^\circ$. These views are randomly sampled without enforcing any preference for frontal views during the training of \NAME{}.

\begin{figure*}[t!]
    \centering
    \includegraphics[width=.9\linewidth]{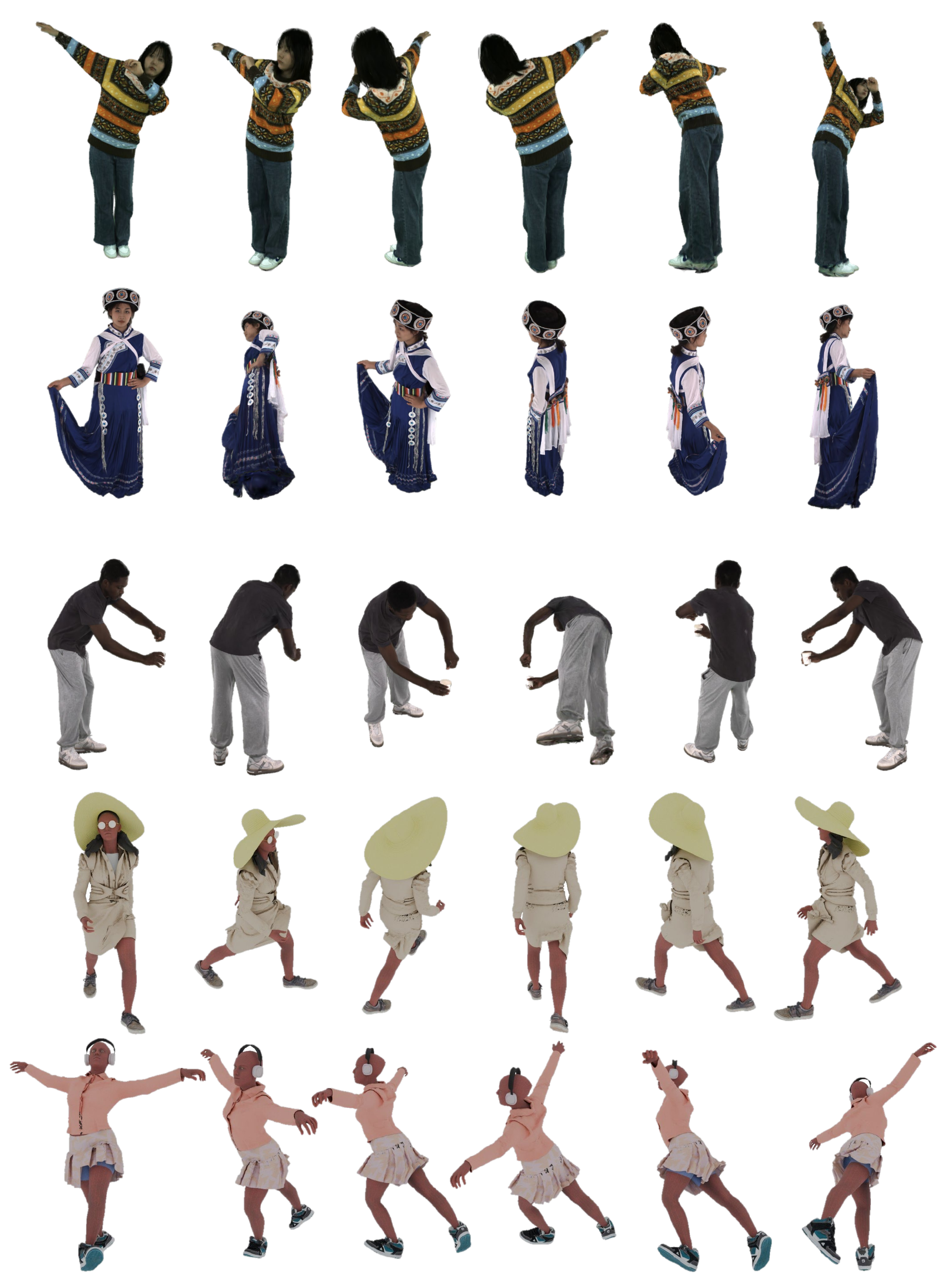}
    \caption{\textbf{Visualization of our training data.} (Best viewed in color.) The first three rows correspond to real-world generated data,
while the remaining rows are generated synthetic data.}
    \label{fig:vis_supp}
\end{figure*}

\noindent \textbf{\NAME{} traning.} The input image resolution of \NAME{} is 512$\times$512 and we crop a 180$\times$180 patch as supervision.
The NeRF~\cite{mildenhall2020nerf} MLP contains 10 layers with the width set as 60 and SiLU~\cite{elfwing2018sigmoid} is utilized as the activation function.
The number of samples per ray is set as 128.
For datasets with 3D scans such as THuman, CustomHuman, and 2K2K, we follow a unified preprocessing pipeline. 
We place each 3D mesh under a canonical camera setup and render 36 multi-view images at 10-degree intervals along a horizontal 360° circle. These rendered RGB images are then used as the supervision signals during training. Importantly, we do not use the original 3D meshes themselves for supervision, but only the rendered images.

\noindent\textbf{Patch selection.} We utilize patches from the image to provide the supervision signals. 
Its selection process is formulated as follows.
Given an image $I \in \mathbb{R}^{H \times W \times C}$ and a corresponding binary mask $M \in \{0, 1\}^{H \times W}$, we aim to sample a square patch of size $r \times r$ (e.g., $r = 180$) such that the region contains a sufficient proportion of foreground pixels.
Let $R_{(i,j)} \subseteq M$ denote a square region of size $r \times r$ with its top-left corner at position $(i, j)$. The foreground ratio of region $R_{(i,j)}$ is computed as:
$$ \alpha_{(i,j)} = \frac{1}{r^2} \sum_{u=0}^{r-1} \sum_{v=0}^{r-1} M_{i+u, j+v}. $$
We define the candidate set of regions as: $ \mathcal{S} = \lbrace (i, j) \mid \alpha_{(i,j)} \geq \tau,\; i, j \in [0, H - r] \; \text{stride} \; s \rbrace $.
Here, $\tau$ is the mask threshold (e.g., $\tau = 0.05$), and $s$ is the stride used to slide the sampling window (e.g., $s = 10$).
For each valid region $(i, j) \in \mathcal{S}$, we assign a sampling weight proportional to the number of foreground pixels:
$$ w_{(i,j)} = \sum_{u=0}^{r-1} \sum_{v=0}^{r-1} M_{i+u, j+v}. $$
The weights are normalized 
into a valid probability distribution:
$$ p_{(i,j)} = \frac{w_{(i,j)}}{\sum_{(i',j') \in \mathcal{S}} w_{(i',j')}}. $$
Finally, a region $(i^*, j^*) \sim p$ is sampled, and the corresponding image patch is extracted as $ P = I[i^* : i^* + r,\; j^* : j^* + r,\; :] $.
This strategy encourages sampling of patches rich in foreground content while preserving diversity.

\begin{figure}[t]
    \centering
    \includegraphics[width=.88\linewidth]{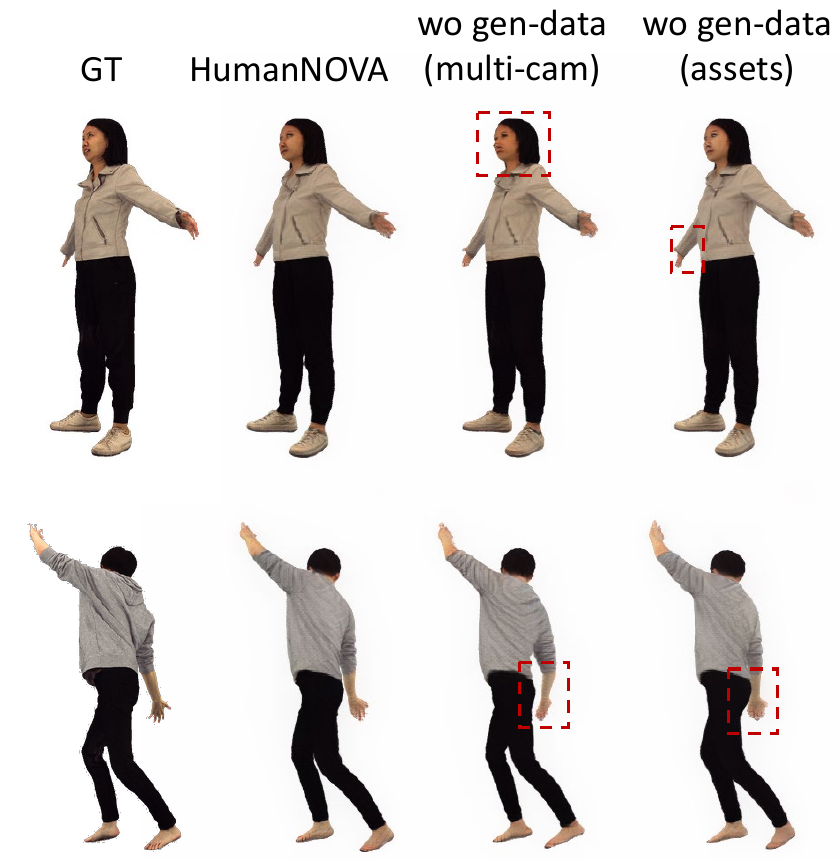}
    \caption{\textbf{Visualization of ablation studies on generated data type.} (Best viewed in color.) Removing the generated data (multi-cam or assets) negatively affects the model performance.}
    \label{fig:ablation1}
\end{figure}

\begin{figure}[t]
    \centering
    \includegraphics[width=\linewidth]{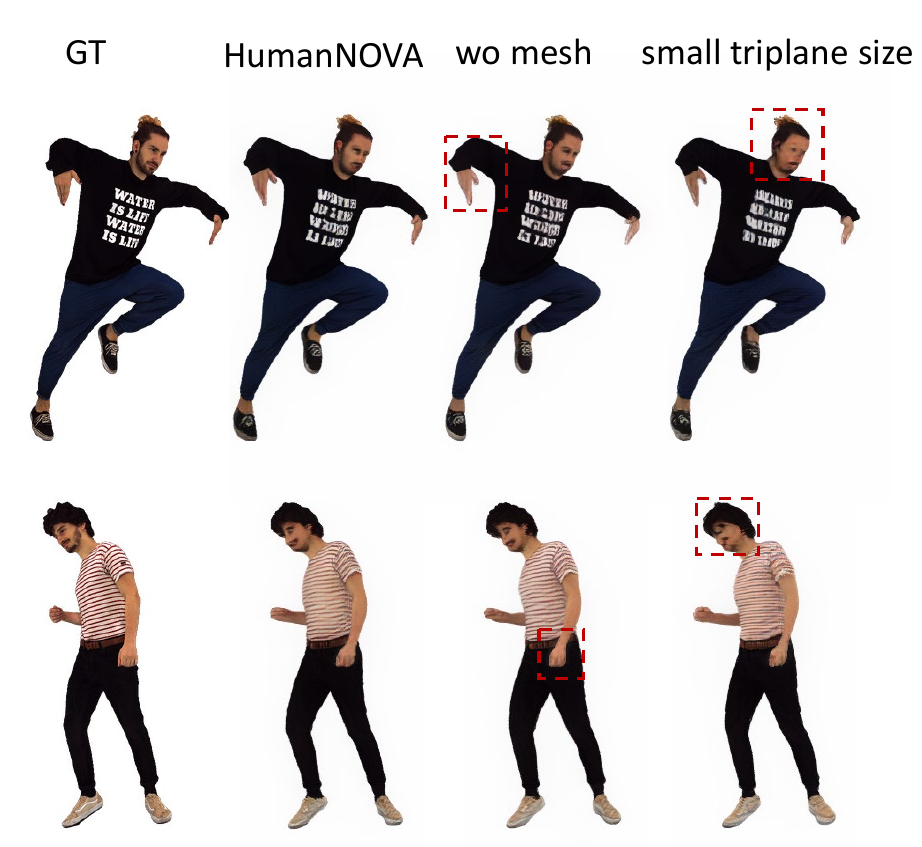}
    \caption{\textbf{Visualization of ablation studies on model design.} (Best viewed in color.) 
    Excluding the mesh prior or reducing the triplane resolution 
    reduces the quality of the output.}
    \label{fig:ablation2}
\end{figure}

\noindent\textbf{Evaluation.} 
We utilize 50 assets from each dataset (CustomHuman, Thuman2, and 2K2K) for evaluation.
Both CustomHuman and THuman2 feature diverse clothing styles, including loose garments and layered outfits, as well as a wide range of body poses, making them particularly challenging for accurate 3D reconstruction.
In contrast, the 2K2K dataset primarily consists of humans in upright, standing poses.
For additional 3D geometry evaluation, we extract the isosurface based on Marching Cubes~\cite{lorensen1998marching} to convert \NAME{}'s implicit representations into meshes.
During evaluation, CD is measured in centimeters (cm), providing a precise indication of surface accuracy. 
F-Score is computed with a threshold of 0.01 meters.

\section*{C. More Visual Results}
\label{sec:visual}

In Figure~\ref{fig:vis_supp}, we visualize a sample from our training data. 
The first three rows correspond to real-world generated data, while the remaining rows are generated synthetic data. 
Together, they provide the foundational training data that empowers \NAME{} to learn robust and generalizable 3D human representations.

We provide visualization of our ablation studies in Figure~\ref{fig:ablation1} and Figure~\ref{fig:ablation2}. 
Figure~\ref{fig:ablation1} illustrates the impact of different training data configurations.
Removing gen-data (multi-cam) leads to less photorealistic results or less accurate structure. 
In contrast, removing gen-data (assets) weakens the model’s capability to perceive and reconstruct human poses (see the right shoulder). 
Figure~\ref{fig:ablation2} shows the effects of varying model settings. 
Excluding the mesh prior degrades the structural quality of the output, while reducing the triplane resolution compromises the model's capacity to represent fine-grained details (see the red box).

To better showcase the results, we include additional in-the-wild reconstruction examples and comparisons with previous methods in the supplementary video.
In the video, the reconstruction results are presented with 360-degree rotation.
From the video, it could be observed that our method is generally not affected by the Janus problem. 
This is because we model the 3D human directly as a whole in 3D space, instead of decomposing the task into separate front/back generation followed by heuristic merging, which is a common cause of inconsistent geometry or appearance across views (e.g., SiTH).